\documentclass[11pt,oneside,a4paper]{article}
\usepackage[utf8]{inputenc}
\usepackage{cite}
\usepackage{amsmath,amssymb,amsfonts}
\usepackage{tabularx}
\usepackage{graphicx}
\usepackage{textcomp}
\usepackage{threeparttable}
\usepackage{tabulary}
\usepackage{color,soul}
\usepackage{authblk}
\usepackage[margin=0.5in]{geometry}
\usepackage{subcaption}
\usepackage{url}
\usepackage{lineno}
\usepackage{float}
\usepackage{soul}
\usepackage{multirow}
\usepackage{tabularx}
\usepackage{rotating}
\usepackage{algpseudocode}
\usepackage{algorithm}
\usepackage[titletoc]{appendix}
\usepackage{mathtools}
\usepackage{amsmath}
\usepackage{rotating}

\usepackage{url}

\title{Ontology in Hybrid Intelligence: a concise literature review}

\author{Salvatore Flavio Pileggi\\\textit{SalvatoreFlavio.Pileggi@uts.edu.au}} 
\affil{University of Technology Sydney\\15 Broadway, Ultimo NSW 2007, Australia}

\date{...}
 
\begin{document}

\begin{titlepage}
\maketitle
\end{titlepage}
 
\begin{abstract}
In a context of constant evolution and proliferation of AI technology,
Hybrid Intelligence is gaining popularity to refer a balanced coexistence between human and artificial intelligence. The term has been extensively used in the past two decades to define models of intelligence involving more than one technology. 
This paper aims to provide (\textit{i}) a concise and focused overview of the adoption of Ontology in the broad context of Hybrid Intelligence regardless of its definition and (\textit{ii}) a critical discussion on the possible role of Ontology to reduce the gap between human and artificial intelligence within hybrid intelligent systems. 
Beside the typical benefits provided by an effective use of ontologies, at a conceptual level, the conducted analysis has pointed out a significant contribution of Ontology to improve quality and accuracy, as well as a more specific role to enable extended interoperability, system engineering and explainable/transparent systems. Additionally, an application-oriented analysis has shown a significant role in present systems (70+\% of the cases) and, potentially, in future systems. However, despite the relatively consistent number of papers on the topic, a proper holistic discussion on the establishment of the next generation of hybrid-intelligent environments with a balanced co-existence of human and artificial intelligence is fundamentally missed in literature. Last but not the least, there is currently a relatively low explicit focus on automatic reasoning and inference in hybrid intelligent systems. 
\end{abstract}


%
%
%
%
%
%
%


\section{Introduction}

Artificial Intelligence is constantly evolving~\cite{muller2016future} towards a more and more sophisticated technology~\cite{lu2019artificial}. It is generating a disruptive innovation in most application domains such as, among others, medicine~\cite{hamet2017artificial} / healthcare~\cite{yu2018artificial}, education~\cite{chen2020artificial}, finance~\cite{bahrammirzaee2010comparative}, transportation~\cite{abduljabbar2019applications} and manufacturing~\cite{li2017applications}. Indeed, the last generation of AI-powered systems is having a clear impact to solve real-world problems, including also tangible contributions in situations of major crisis~\cite{vaishya2020artificial}, and, in general terms, is generating probably unprecedented expectations.

The current socio-technological climate is characterised by a imperishable mix of enthusiasm and concern. A concrete example is generative AI~\cite{walters2020assessing}, such as ChatGTP, an AI program that has become a cultural sensation in less than 2 months~\cite{thorp2023chatgpt}. By automatically creating text based on written prompts, ChatGTP has revealed the potentiality of AI also to non-expert users, while has further increased concerns on possible and probable mis-use, as well as on collateral effects for the society. The classic debate involving experts from academia and industry around ethics and possible regulations, is now somehow extended to involve a much broader audience in a kind of public debate.

The rising of AI has also led to an increasing popularity of hybrid solutions, normally referred to as Hybrid Intelligence~\cite{dellermann2019hybrid}.
That is an evidently less radical concept that assumes a model of intelligence based on a balanced co-existence of humans and machines. To remark that the concept has been extensively used in the past two decades to define models of intelligence involving more than one technology. 

Born in philosophy, Ontology is a relatively consolidated concept also in Computer Science~\cite{guarino1995formal}, where it is extensively adopted to formalize a given domain within computer systems~\cite{chandrasekaran1999ontologies}. In practice, Ontologies are rich data models that allow the specification of machine-processable "semantics". While their primary purpose is machine-to-machine interaction, ontological modelling enables extended capabilities also in terms of human-computer interaction by fostering the same representation of knowledge for humans and machines. While ontology actively contributes in general in the context of Intelligent Systems at a different level, its specific potential within Hybrid Intelligence is probably not fully exploited and could play a significant role within the next generation systems.  

In a context of constant evolution and proliferation of AI technology, this paper aims to provide (\textit{i}) a concise and focused overview of the adoption of Ontology in the broad context of Hybrid Intelligence regardless of its definition and (\textit{ii}) a critical discussion on the possible role of Ontology to reduce the gap between human and artificial intelligence. 

The relevance of this study resides mostly in a need for a consolidated approach for Hybrid Intelligence that may assure an evolution according to the current society needs. 

\paragraph{Why Hybrid Intelligence?}
Regardless of concrete implementations, hybrid solutions are normally designed to take advantage of the characteristics of multiple approaches/technologies or to optimise some kind of key trade-off. In a context of huge proliferation of HI, Hybrid Intelligence is mostly emerging as a result of two possible main paths: (i) \textit{current limitations of AI technology} and/or (ii) \textit{lack of trust in fully AI-based solutions}. While the relevance of the former trend is naturally expected to decline with the evolution of AI, a more sophisticated AI technology could have an overall unclear impact on the latter as a more powerful AI could further increase the concern for sensitive applications.

\paragraph{Why Ontology?}
Ontology may be considered a consolidated asset within the Semantic Web~\cite{berners2001semantic}. It is contributing more in general to solve problems in Computer Science (e.g. data integration~\cite{noy2004semantic}) and is extensively adopted in a broad range of disciplines and application domains (e.g. bio-informatics)~\cite{stevens2000ontology}, as well as it is object of constant research interest, as specialised journals and conferences clearly demonstrate. Despite its already mentioned consolidated value within computer systems, Ontology is evolving as a response to emerging problems and major technology trends. For instance, it is remarkable the convergence with Machine Learning (e.g.~\cite{sacha2018vis4ml}, among the very many) and its intrinsic relation with Knowledge Graphs~\cite{pujara2013knowledge}, as well as the contribution at an application level to emerging systems (e.g. Digital Twins\cite{karabulut2023ontologies}).

\paragraph{Structure of the paper.} {The introductory part follows with a brief presentation of the background concepts (Section~\ref{sec:background}); methodological aspects are discussed in Section~\ref{sec:methodology}, while Section~\ref{sec:review} provides an overview of the contributions on the topic in literature; such a body of knowledge is discussed in quantitative and qualitative terms in Section~\ref{sec:discussion}; finally, the paper ends with conclusions section.}

\section{Background concepts}\label{sec:background}

The root of this review is intrinsically related to the very generic concepts of \textit{intelligence} and \textit{semantics}, which have an extremely broad scope in the context of modern systems. The focus is explicitly on more narrowed concepts, more concretely on the intersection between \textit{Hybrid Intelligence} and \textit{Ontology}. 
Such an approach allows the identification of a body of knowledge. However, hybrid intelligent systems may be defined in a different ways and quite often present blurred boundaries; furthermore, ontologies may refer to the more generic and conceptual knowledge representation as well as to the more technology-specific Semantic Web. 

While methodological aspects are proposed later on (Section~\ref{sec:methodology}), this section aims to overview the central concepts in context. An in-depth exhaustive discussion is out of the paper scope.  

\subsection{Defining "intelligence"}

The definition of "intelligence" has been historically object of discussion and controversy. While it is not probably possible to provide universally accepted formal definitions, it is interesting to look at the understanding of intelligence in the context of a more and more technology-intensive society.

\subsubsection{Human and Artificial Intelligence}
Human (or natural) intelligence is normally associated with the ability to generate an abstracted mental model of a given reality and perform mental simulations accordingly. It underlays the common notions of thinking and reasoning~\cite{ma2011cyber} enabling human skills in terms of problem analysis/solving and decision making~\cite{klein2017sources}. 

Although such considerations may be considered a common assumption, it is also well-known that human intelligence goes beyond its "analytical" component to embrace also "emotional" intelligence~\cite{mayer2008human}, which is based on emotions (e.g. empathy) to enable a unique and strictly human way to behave, reason and decide resulting from a mix of rational elements and "gut feelings"~\cite{klein2004power}. This more comprehensive understanding of intelligence plays a key role given the increasing dehumanization within society, which is pushed also by technology~\cite{oviatt2021technology}.

There are at least two critical complementing considerations: on one side, all human beings have their own consciousness and unique ways of thinking~\cite{ma2011cyber}; on the other side, individuals belong to a social context, which affects the way reality is perceived and consequently the way to think. It naturally leads to the concept of "collective" intelligence~\cite{leimeister2010collective}, which puts emphasis on the added, if not determinant, value of the collectivity to establish a level of intelligence higher than the correspondent to any individual. Collective intelligence becomes even more critical in the context of an always connected world with unprecedented interaction possibilities enabled by the evolution of the Web~\cite{o2005web} and social Web sites operating at a global scale~\cite{kim2010social}.

Computers have intrinsically generated the idea of machines able to "reason" like humans. Indeed, Alan Turing, often considered the "father of computer science" formulated the question "can machines think?" that led to the famous "Turing Test", where a human tries to distinguish between a computer and human response in a conversation~\cite{machinery1950computing}. Since then, the intuitive concept of Artificial Intelligence has evolved in the scientific context and very many definitions have been provided. For instance,
"AI is the science and engineering of making intelligent machines, especially intelligent computer programs"~\cite{mccarthy2007artificial}.

Given the huge availability of data and computational resources, the most recent advances in the field are producing a actual new generation of solutions, which are often converted to services or tools potentially available to the most, such as the already cited ChatGPT. Main stakeholders are actively involved in a constant debate, with a mix of excitement and concern. It is triggering different feeling and perceptions in the broader social context.

\subsubsection{Hybrid Intelligence}

As part of the ongoing debate, more and more questions are rising up. Among them, and because of a number of evident concerns, 
\textit{how human and artificial intelligence are going to co-exist?}; \textit{how such a "relationship" is going to evolve in the future?}

A an interesting (and also very intuitive) concept in this broader context is probably Hybrid Intelligence, that focuses on expanding the human intellect with AI, instead of replacing it~\cite{ho2022argumentation}. Authors in \cite{dellermann2019hybrid} characterise Hybrid Intelligence through a comparative analysis.

This more recent understanding and definition of Hybrid Intelligence as a combination of human and machine intelligence also reflects an evolution of the concept. Indeed, that same term has been extensively adopted more or less informally in the last two decades to indicate the engineering of systems whose "intelligence" results from the combined application of more than one technology.

\subsection{Semantics, Ontology and Semantic Web}

Born as the branch of metaphysics dealing with "the nature of being",  ontology is a philosophical concept often associated with a formal conceptualization of a given domain. 
According to that same holistic approach, ontology became part of the computer world~\cite{guarino1995formal} to provide machine-processable conceptualizations. 

An ontology-based approach to Knowledge Representation allows to specify "semantics" within computer systems and, indeed, ontology is a key asset within the Semantic Web~\cite{berners2001semantic}. Semantic Web adopts the Web infrastructure and standard languages to formalise a common understanding of data and to achieve "intelligence" via interoperability and inference. Ontologies are extensively adopted in wide range of domains and applications, as well as they play a role in establishing advanced data ecosystems, such as Open Data~\cite{murray2008open} and Linked Data~\cite{bizer2011linked}.

The primary application context for ontology is machine-to-machine interaction. However, through a proper knowledge  engineering process, it is possible to assume a "shared" vision of a given knowledge for humans and machines. It can be further fostered by different kind of abstractions and visualizations~\cite{pileggi2022getting}.   

\section{Methodology and approach}\label{sec:methodology}

\begin{figure}[!h]
\centering

\begin{subfigure}[b]{\textwidth}
\includegraphics{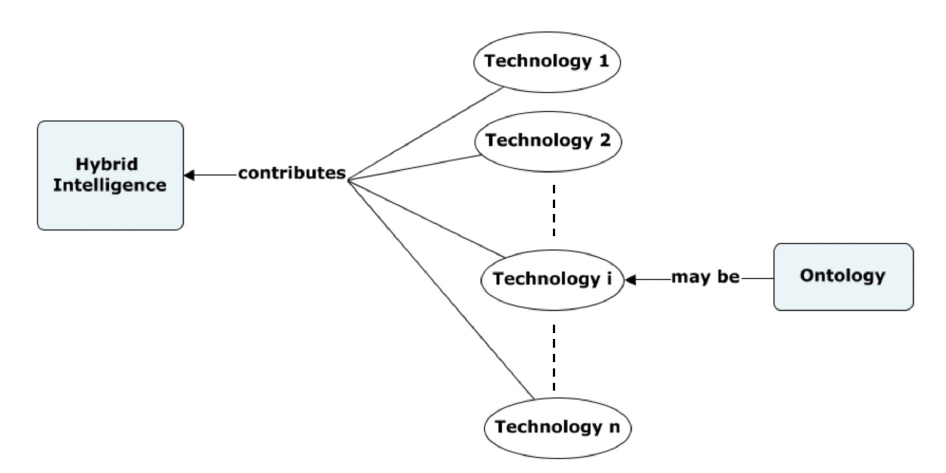}
\caption{Historical common understanding of HI and related Ontology role.}\label{fig:HI-historical}
\end{subfigure}

\begin{subfigure}[b]{\textwidth}
\includegraphics{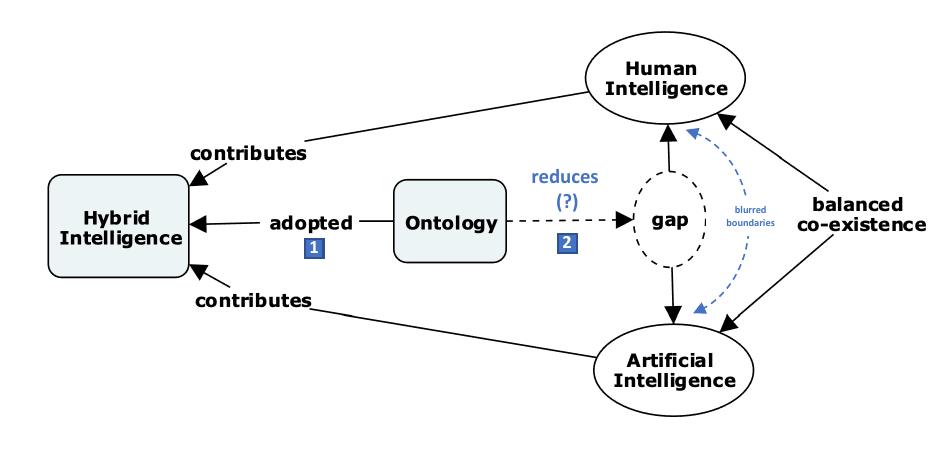}
\caption{HI evolution with the proliferation of AI technology and related Ontology role.}\label{fig:objectives}
\end{subfigure}

\caption{Conceptualization of main objectives and scope of the review through the analysis of the HI concept evolution.}
\end{figure}

This paper aims to provide (\textit{i}) a concise and focused overview of the adoption of Ontology in the broad context of Hybrid Intelligence and (\textit{ii}) a critical discussion on the possible role of Ontology to reduce the gap between human and artificial intelligence within hybrid intelligent systems. A simplified conceptualization of main objectives and scope by adopting an enriched conceptual map \cite{canas2005concept} is proposed in Figure ~\ref{fig:objectives}. The underlying idea assumes intelligence enabled in a heterogeneous context, either involving multiple technologies or a coexistence of human and artificial intelligence.   

Because of the potentially broad scope, looking at that specific context, the literature review has been conducted assuming the following selection criteria:
\begin{description}
\item[SC.1] Selected papers have an explicit focus on Hybrid Intelligence, regardless of its contextual definition
\item[SC.2] Selected papers explicitly address the role of Ontology
\item[SC.3] The role of ontology is relevant in the context of the contribution and its value can be identified
\end{description}

While searching criteria have been designed accordingly to systematically retrieve papers which simultaneous contain the keywords "\textit{Hybrid Intelligence}" and "\textit{Ontology}", acceptance is based on a qualitative assessment by the author of \textit{SC.1}, \textit{SC.2} and \textit{SC.3}. However, as ontology and hybrid intelligence can be indirectly referred in different ways within the current technological context, the query has been extended to include possible common equivalent/strictly related  concepts as follows:

\begin{verbatim}
("Hybrid Intelligence" OR "Hybrid Intelligent") AND ("Ontology" OR "Semantic Web")
\end{verbatim}

On the other side, very generic keywords, such as "Knowledge Representation" have not been used for retrieval.

In order to further enhance a critical discussion in context, selected papers, wherever possible, are associated with a generic domain and even with a more specific application. Additionally, as part of the adopted framework, the different contributions are classified depending on their focus (conceptual/theoretical or application-focused) and on the provided value in a given system. This light classification model contributes to generate a structured overview and a conceptualised analysis. On the other side, classifications are arbitrary and may present some overlapping.    

The combined application of the mentioned selection criteria results in a relatively restrictive retrieval strategy that limits significantly the number of papers to include in the literature review. In this specific case, this approach assures a focused discussion within a broad topic, mostly characterised by the blurred boundaries of Hybrid Intelligence.

\section{Ontology in Hybrid Intelligence}\label{sec:review}

An overview of the major contributions retrieved from the most common scientific databases is proposed in Table~\ref{table:conceptualContributions} and~\ref{table:applicationContributions}, which report conceptual and application-focused papers respectively. In order to provide a grounded discussion, they are addressed separately in the next two sub-sections.

\subsection{A conceptual perspective}

Table~\ref{table:conceptualContributions} reports the contributions in literature that present a more theoretical/conceptual approach. Although they have not a specific application-focus and should allow a certain generalization of the insight, where specified, the original domain/application is still reported. 

\begin{table}[!h]
\small
\centering
\begin{tabular}{ |c|c|c|c|c| } 
\hline
\textbf{Contr.} & \textbf{Domain} & \textbf{Application} & \textbf{Focus} & \textbf{Value} (main)\\
\hline
\hline
\cite{dellermann2019design} & Business & Decision Support System & Conceptual & Interoperability\\
\cite{krinkin2022co} & N\textbackslash A & N\textbackslash A & Conceptual & Interoperability\\
\cite{krinkin2023cognitive} &N\textbackslash A & N\textbackslash A & Conceptual & Interoperability\\
\cite{shichkina2022principles} & Healthcare & Explainable models & Conceptual & Explainability\\
\cite{ho2022argumentation}  &N\textbackslash A & Explainable models & Conceptual & Explainability\\
\cite{bredeweg2022requirements} & Education & System Thinking & Conceptual & System Engineering \\
\cite{garcia2013ontological} & Smart Systems & Ambient Assisting Living & Conceptual & System Engineering \\ 
\cite{rodriguez2021introduction} & N\textbackslash A & N\textbackslash A & Conceptual & Explainability\\
\cite{pan2016heading} & N\textbackslash A & Collective Intelligence & Conceptual & Quality and Accuracy\\
\cite{tiddi2022knowledge} & N\textbackslash A & Knowledge Graph & Conceptual & Explainability\\
\cite{moradi2019search} & N\textbackslash A & Collective Intelligence & Conceptual & Quality and Accuracy\\
\cite{zheng2017hybrid} & N\textbackslash A & N\textbackslash A & Conceptual & Quality and Accuracy\\
\cite{van2021modular} & N\textbackslash A & N\textbackslash A & Conceptual & System Engineering\\
\cite{shunkevich2022ontology} & N\textbackslash A & N\textbackslash A & Conceptual & System Engineering\\
\hline
\hline
\end{tabular}
\caption{\label{table:conceptualContributions}Overview of the selected contributions with a conceptual focus. }
\end{table}

In~\cite{dellermann2019design}, authors propose a design principle: "\textit{Provide the Hybrid Intelligence DSS with an ontology-based representation to transfer an entrepreneur’s assumptions and create a shared understanding among the mentors, the machine, and the entrepreneur}". Despite the domain/application specific focus, the use of ontologies to create a shared understanding among all stakeholders (including also machines) can be applied in general terms.

Another principle assumes interoperability as one of the key factors behind the development of hybrid intelligence~\cite{krinkin2022co}. An effective machine-to-machine model to assure interoperability among different systems leads to a central role of ontologies to enable sophisticated approaches (Semantic Interoperability~\cite{obrst2007evaluation}).
Additionally, also for interoperability, a multi-stakeholder vision fundamentally applies to real-world scenarios that typically involve machine both with designers, subject matter experts and final users~\cite{krinkin2023cognitive}.

Looking at the current technological climate, an aspect of increasing interest is the potential role of ontology to reduce fear of AI by pushing more intelligible explainable models
~\cite{shichkina2022principles}\cite{rodriguez2021introduction}. It would enable more transparent and trusting environments for an effective co-existence of human and artificial intelligence, especially if considered both with Knowledge Graphs~\cite{tiddi2022knowledge}. Moreover, a wider perspective to possible evolutions of AI that implies an increasing convergence of data, such as AI 2.0~\cite{pan2016heading}, is worthy of mention. 

A relatively more classic contribution is in the engineering of systems adopting hybrid intelligence, where ontology can contribute to formalize the specification of requirements \cite{garcia2013ontological}. This is in line with typical benefits of Ontology in the generic System~\cite{yang2019ontology} / Software~\cite{pileggi2018ontology} Engineering, as well as in the more specific field of Requirement Engineering~\cite{dermeval2016applications}. Given the recent advancement in AI and the peculiarities of Hybrid Intelligence, the role of Ontology throughout the design process could be remarkable (e.g. contributing to formalization of problems~\cite{shunkevich2022ontology}) and can eventually provide also linking to the different principles, such as ethics. In~\cite{bredeweg2022requirements}, authors further extend the intent and the extent within system engineering by dealing with requirements and challenges. It addresses the need for automatic reason.  

Last but not least, ontologies may contribute to enable and sustain the quality of knowledge-based systems. For instance, argumentation can be adopted to address inconsistencies in knowledge bases~\cite{ho2022argumentation}.

\cite{moradi2019search} deals with the interesting concept of Hybrid Collective Intelligence and the consequent need for knowledge convergence, as well as \cite{zheng2017hybrid} discusses the augmentation of Hybrid Intelligence by collaboration and cognition with an extensive use of semantic artefacts. In general terms, ontology-based models are extensively used in hybrid learning~\cite{van2021modular}. 

\subsection{An application perspective}

An overview of applications is summarised in Table~\ref{table:applicationContributions}. Such an application perspective shows a clear generic focus on Knowledge Representation.

\begin{table}[!h]
\small
\begin{tabular}{ |c|c|c|c|c| } 
\hline
\textbf{Contr.} & \textbf{Domain} & \textbf{Application} & \textbf{Focus} & \textbf{Value} (main)\\
\hline
\hline
\cite{fill2020supporting} & N\textbackslash A & Blockchain & Application & Automatic Reasoning\\
\cite{watanobe2014hybrid} & Software & Programming & Application & Knowledge Representation\\
\cite{palvannan2023hias} & Agriculture & Recommender Systems & Application & Semantic Similarity\\
\cite{isaza2009towards} & Security & Intrusion Detection/ Prevention & Application & Modelling\\
\cite{hingant2018hybint} & Security & Critical Infrastructure Protection & Application & Information Standardization\\
\cite{hwang2003hybrid} & Smart Systems & Assistive Technology & Application & Knowledge Representation\\
\cite{lee2022stargazer} & Healthcare & Drug Discovery & Application & Analysis\\
\cite{michaelis2015explaining} & Management & Decision Support Systems & Application & Knowledge Representation\\
\cite{reitemeyer2020automatic} & Business & Automated Enterprise Modeling & Application & Automated Reasoning \\
\cite{levy2021assessing} & Healthcare & Decision Support Systems & Application & Knowledge Representation\\
\cite{palagin2017noosphere} & Science & N\textbackslash A  & Application & Knowledge Representation\\
\cite{pankowski2011combining} & N\textbackslash A & Data Integration & Application & Knowledge Representation\\
\cite{graef2021human} & Business & Assistive Technology & Application & Semantic Similarity\\
\cite{taran2021text} & Law & Text mining & Application & Knowledge Representation\\
\cite{chernenkiy2018hybrid} & N\textbackslash A & Information Systems & Application & Knowledge Representation\\
\cite{cheng2014social} & N\textbackslash A & Internet of Things & Application & Knowledge Representation\\
\cite{listopad2020similarity} & N\textbackslash A & Multi-agent Systems & Application & Semantic Similarity \\
\cite{sarwar2022context} & N\textbackslash A & Data Analysis & Application & Knowledge Representation\\
\cite{belhadi2022hybrid} & Healthcare & Automated Learning & Application & Knowledge Representation\\
\cite{kirikov2021agents} & N\textbackslash A & Multi-agent Systems & Application & Automated Reasoning\\
\cite{hadjiski2007hvac} & Smart Systems & Control Systems & Application & Knowledge Representation\\
\cite{listopad2020estimating} & N\textbackslash A & Multi-agent Systems & Application & Semantic Similarity\\
\cite{listopad2020cohesive} & N\textbackslash A & Multi-agent Systems & Application & Knowledge Representation\\
\cite{zhang2002agent} & Business & Multi-agent Systems & Application & Knowledge Representation\\
\cite{listopad2020modeling} & N\textbackslash A & Multi-agent Systems & Application & Knowledge Representation\\
\cite{bravo2005ontology} & N\textbackslash A & Multi-agent Systems & Application & Knowledge Representation\\
\cite{rumovskaya2020visualization} & N\textbackslash A & Multi-agent Systems & Application & Knowledge Representation\\
\cite{hadjiski2016integration} & N\textbackslash A & Multi-agent Systems & Application & Automated Reasoning\\
\cite{garrido2008using} & N\textbackslash A & Decision Support Systems & Application & Knowledge re-use and Sharing\\
\cite{charest2006ontology} & N\textbackslash A & Data Mining & Application & Automated Reasoning\\
\cite{torshizi2014hybrid} & Healthcare & Recommender Systems & Application & Modelling\\
\cite{belyanova2020using} & N\textbackslash A & Information Systems & Application & Knowledge Representation\\
\cite{hernandez2019hybrid} & Agriculture & Control Systems & Application & Knowledge Representation\\
\cite{varela2020ontology} & N\textbackslash A & Collaborative Systems & Application & Knowledge Representation\\
\cite{li2010agent}  & Business & Recommender Systems & Application & Knowledge Representation\\
\cite{eltaher2008generic} & Software & Software Engineering & Application & Automated Reasoning\\
\cite{zhong2004constructing} & N\textbackslash A & Data Mining & Application & Knowledge Representation\\
\cite{martin2017efficient} & Manufacturing & Industry 4.0 & Application & Knowledge Representation\\
\cite{de2020fog} & Smart Systems & Energy Systems & Application & Knowledge Representation\\
\cite{gouveia2020dss} & N\textbackslash A &Information  Systems& Application & Knowledge Representation\\
\cite{casal2023design} & Management & Decision Support Systems &Application & Knowledge Representation\\
\cite{krieger2008hybrid} & Business & Business Intelligence & Application & Automated Reasoning \\
\hline
\hline
\end{tabular}
\caption{\label{table:applicationContributions}Overview of the selected contributions with an application focus. }
\end{table}

Indeed, Knowledge Representation is seen as an essential asset to trace the provenance of data and knowledge and to link the accountability of humans and machine-based agents~\cite{fill2020supporting}. Such kind of integrated approach is a major prerequisite to establish trust in hybrid systems~\cite{fill2020supporting}. These generic considerations apply in a different way within concrete application domains (e.g. Science~\cite{palagin2017noosphere}). For instance, in certain programming environments based on semantically enriched compound objects, "the confidence of users in making decisions is based on access to previous decisions of others, on quick perception and understanding these decisions and on their transformations leading to new models and algorithms"~\cite{watanobe2014hybrid}. Additionally, the adoption of shared ontologies for different purposes - such as upper, domain, communication and state annotation - is a relatively established pattern~\cite{hwang2003hybrid}.

Within the  broad Knowledge representation context, it is possible to define a set of more fine-grained capabilities. For instance, ~\cite{palvannan2023hias} proposes a model based on Semantic Similarity to generate a knowledge base in the field of Agriculture from multiple sources. Solutions for agriculture are also proposed in~\cite{hernandez2019hybrid}. Semantic Similarity~\cite{listopad2020similarity}\cite{listopad2020estimating}  and negotiation techniques~\cite{kirikov2021agents}\cite{bravo2005ontology} are considered in multi-agent systems. More in general, multi-agent systems extensively adopt ontology-based representations~\cite{listopad2020cohesive}\cite{zhang2002agent}\cite{listopad2020modeling}\cite{rumovskaya2020visualization}.

\cite{isaza2009towards} adopts an ontological model for intrusion detection. The work proposed in~\cite{hingant2018hybint} addresses the significant issues related to information standardization and ontology-based representation in the security domain, while in~\cite{lee2022stargazer} the emphasis is on analysis.

In general terms, automation plays a key role in modern systems. For instance,~\cite{reitemeyer2020automatic} proposes an approach for automated Enterprise Modeling,~\cite{hadjiski2016integration} presents a technique for component integration,~\cite{pankowski2011combining} approaches problems related to metadata management in data exchange
and integration, while~\cite{levy2021assessing} deals with automated  annotation and ontological mapping of clinical texts. Additionally, the system proposed in~\cite{hadjiski2007hvac} has an explicit focus on automated control, as well as~\cite{charest2006ontology} and~\cite{zhong2004constructing} addresses hybrid-intelligent data mining.

In line with the more theoretical analysis previously proposed, the work presented in~\cite{michaelis2015explaining} focuses on the explanation of hybrid intelligence systems with an extensive use of Semantic Web technology, as well as the value of semantics in human-machine collaboration is highlighted in~\cite{graef2021human}. Semantic Web technology is adopted to support hybrid systems in the Law domain ~\cite{taran2021text},  within more generic Information Systems~\cite{chernenkiy2018hybrid}\cite{belyanova2020using}\cite{gouveia2020dss} and Internet of Things~\cite{cheng2014social}. 

The system proposed in~\cite{sarwar2022context} adopts ontology to generate and classify vehicle driver profiles. Ontology matching is a central concept in the automated medical learning approach proposed in~\cite{belhadi2022hybrid}. 

\cite{garrido2008using} implicitly refers to Hybrid Intelligence and addresses the relevance of knowledge re-use and sharing. 

Fuzzy-logic and ontological modelling are combined in~\cite{torshizi2014hybrid} to implement sophisticated recommendation in healthacare. Solutions for Recommendation Systems are also object of other works (e.g.~\cite{li2010agent}). 

Ontology-based meta-modelling is adopted in~\cite{varela2020ontology} to support machine-to-machine as well as human-machine interaction.
Remarkable also applications in the area of Software Engineering~\cite{eltaher2008generic} and advanced manufacturing (e.g. Industry 4.0~\cite{martin2017efficient}), Energy Systems~\cite{de2020fog}, crisis management/prevention~\cite{casal2023design} and Business Intelligence~\cite{krieger2008hybrid}.

\section{Gap Identification and Challenges}\label{sec:discussion}

This section aims to quantitatively and qualitatively analyse the identified body of knowledge and to critically discuss major gaps, with an outlook on challenges.

\subsection{A quantitative analysis}\label{sec:quantitative}

The proposed analysis has been conducted on 56 different papers on the topic. As expected, most contributions (75\%) present an application focus. However, the relatively consistent amount of more conceptual papers has provided a significant insight (Section~\ref{sec:qualitative}). 

\begin{figure}[!h]
\centering
\begin{minipage}{.5\textwidth}
  \centering
  \includegraphics[width=\linewidth]{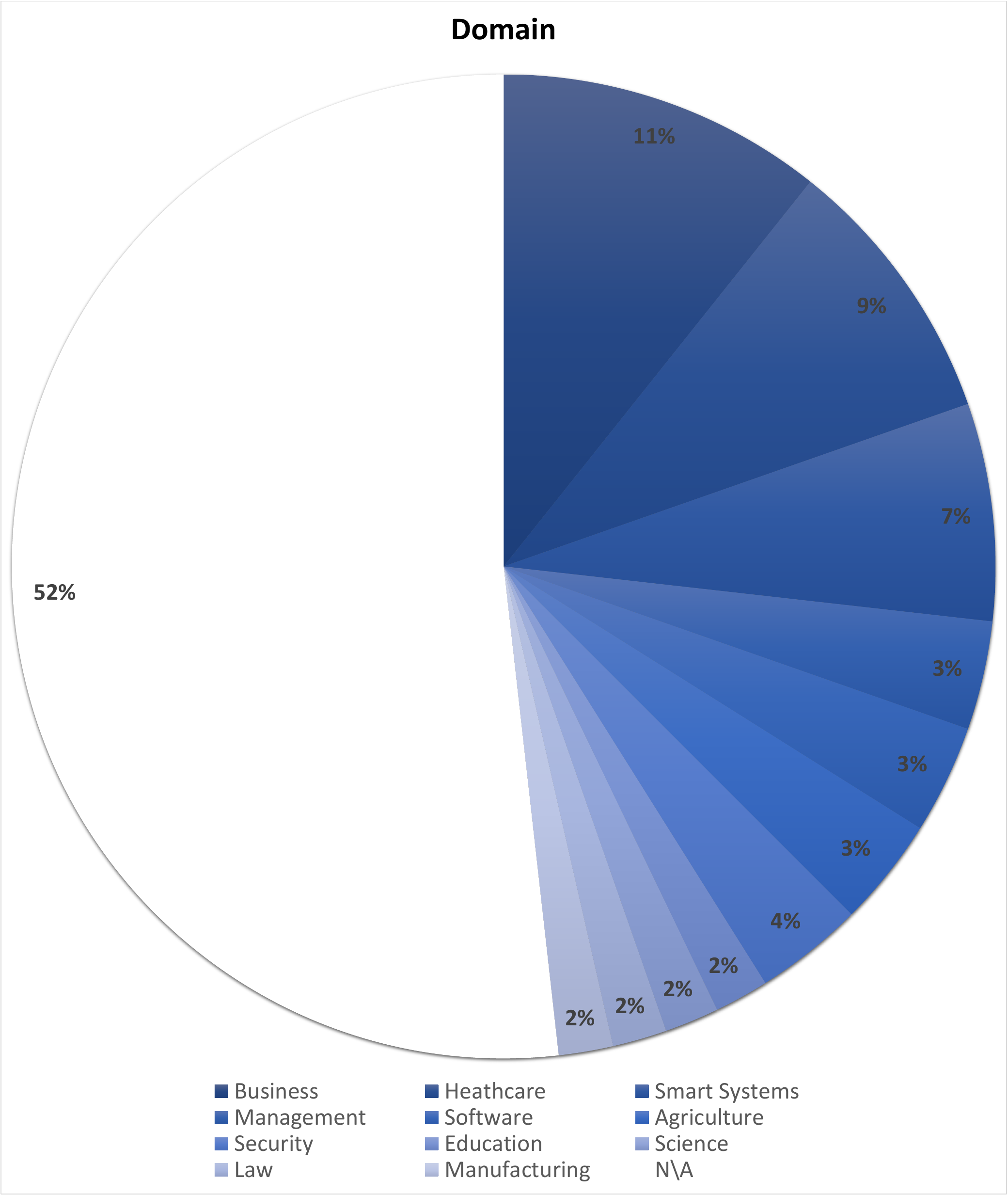}
  \captionof{figure}{Quantitative analysis by macro-domain.}
  \label{fig:domain}
\end{minipage}%
\begin{minipage}{.5\textwidth}
  \centering
  \includegraphics[width=.955\linewidth]{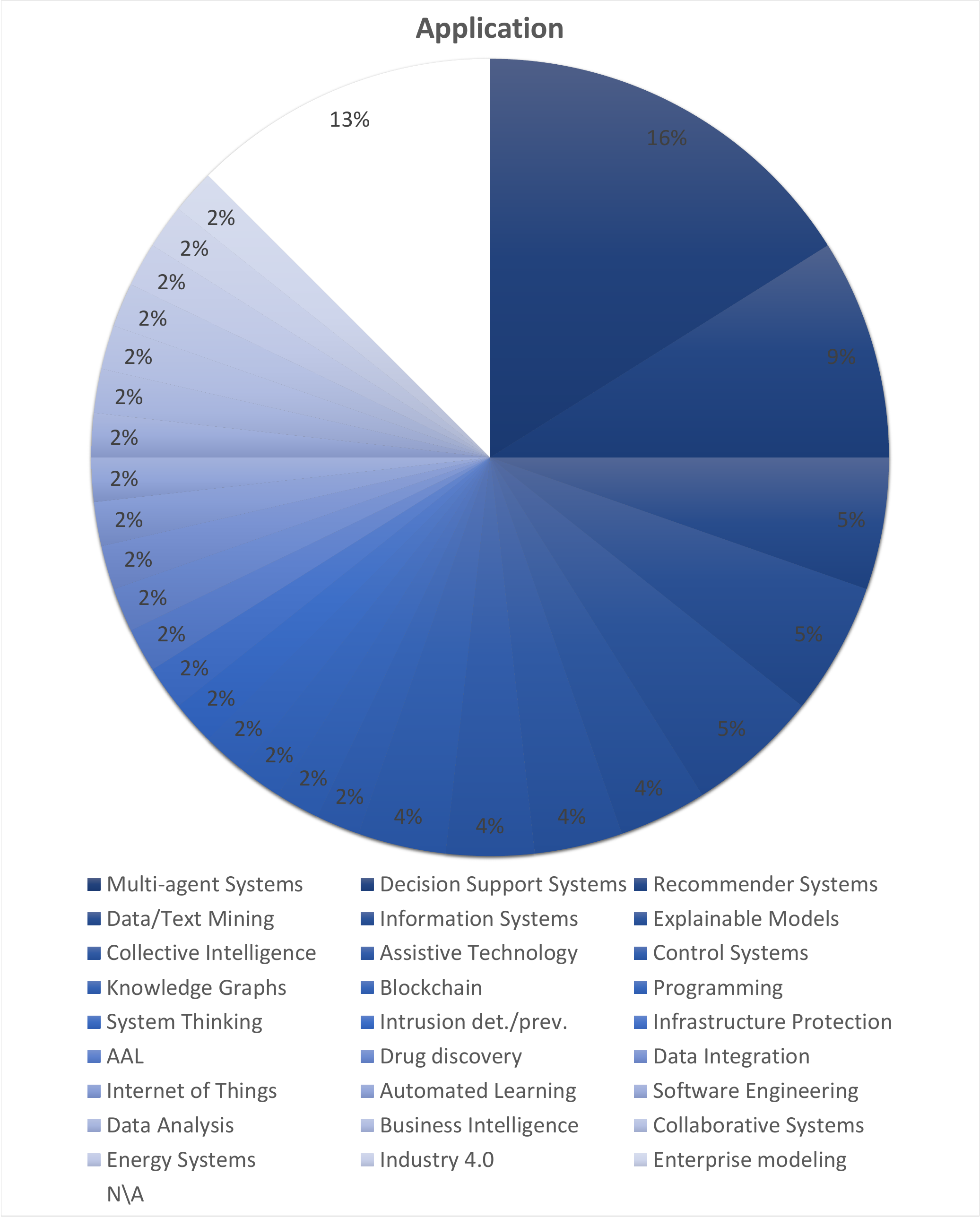}
  \captionof{figure}{Quantitative analysis by application.}
  \label{fig:application}
\end{minipage}
\end{figure}

The classification by macro domain (Figure~\ref{fig:domain}) is relatively  distributed and does not allow the identification of major patterns or trends. Almost an half of the papers cannot be clearly associated with a macro-domain. There is a prevalence of contributions in business (11\%), healthcare (9\%) and smart systems (7\%).

Similar considerations apply also to a quantitative analysis by application (Figure~\ref{fig:application}), which presents a low percentage (13\%) of unclassified papers and shows a major number of contributions in Multi-agent Systems (16\%) and Decision Support Systems (9\%).

\begin{figure}[!h]
\centering
 \centering
  \includegraphics[width=\linewidth]{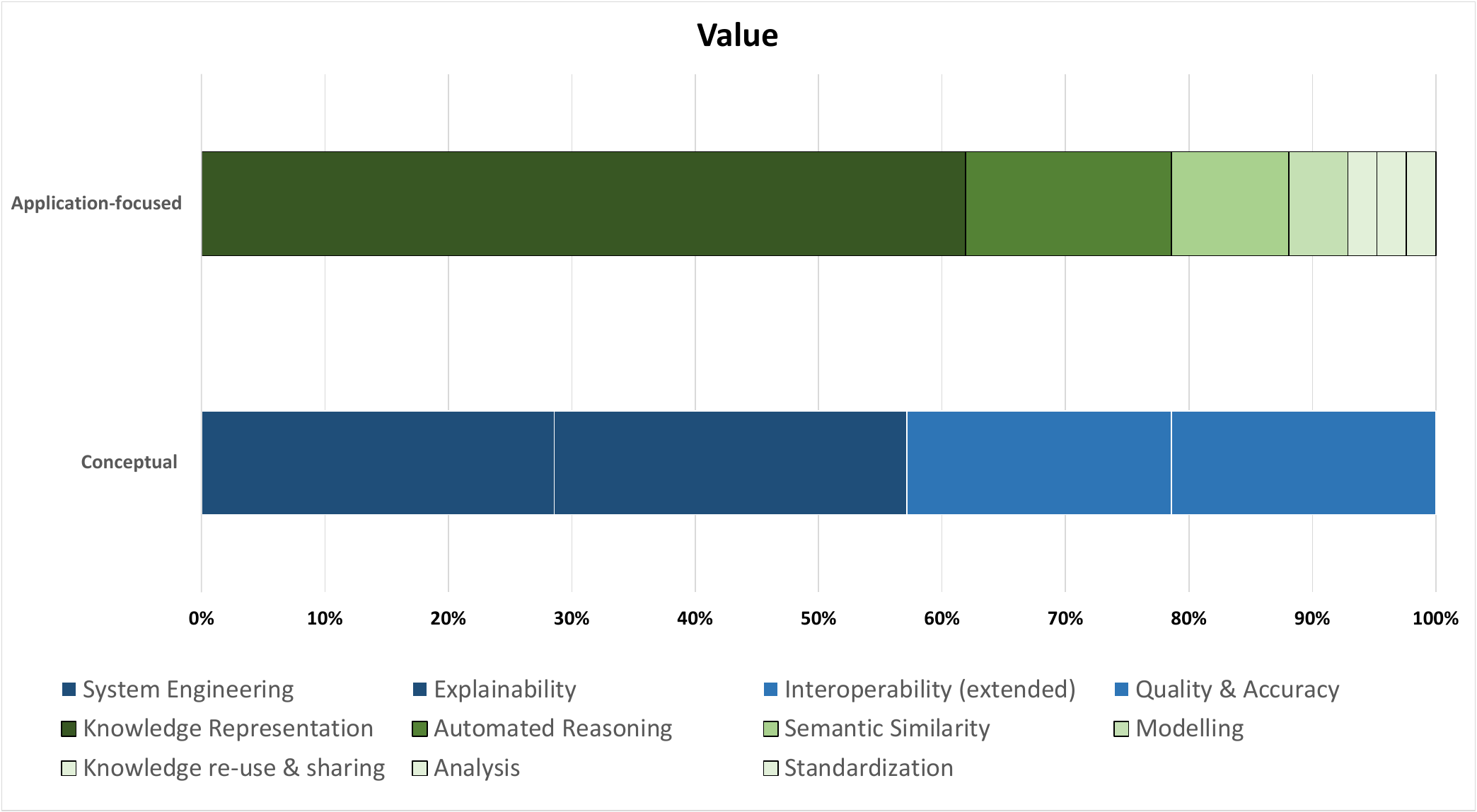}
  \caption{Quantitative analysis of the provided value from a conceptual and an application perspective.} \label{fig:value}.
\end{figure}

An analysis by "provided value" is reported in Figure~\ref{fig:value}. In quantitative terms, the papers object of analysis deal with the four different identified categories uniformly. Such contributions are discussed in qualitative terms later on in the paper. An application-oriented perspective points out a high number of works on generic knowledge representation, while there is much less explicit focus on other aspects, including also automatic reasoning. 

\subsection{A qualitative analysis}\label{sec:qualitative}

From a conceptual point of view, four key aspects have been identified as follows: 

\begin{description}

\item[CP.1]\textit{Interoperability as a key factor}. It includes interoperability among systems as well as a shared understanding between humans and machines. Interoperability is a concept classically associated with Ontology. However, in the specific field of Hybrid Intelligence, there is a clear emphasis on human-machine. It may have implications on Knowledge Engineering, as well as on the evolution of ontology-related technology.

\item[CP.2]\textit{Explainable and transparent models}. The use of ontologies to support some kind of intelligible synthesis of a given analysis or process is not an absolute novelty. For instance, Knowledge Graphs~\cite{hogan2021knowledge} are extensively adopted in several disciplines and applications and may be underpinned by formal ontologies to deal with the underlying complexity~\cite{pileggi2022getting}. In most cases, such a presentation level is ad-hoc designed to optimally match the requirements within a system. Explainable and transparent models in AI (and Hybrid AI) require probably a more systematic approach.

\item[CP.3]\textit{System Engineering}. Also the contribution of Ontology within System/Software Engineering is well known and extensively documented in literature. This review, conducted in the specific context of Hybrid Intelligence, has pointed out a potential extended and enhanced scope to  include linking to different aspects, such as design and ethic principles, and challenges.

\item[CP.4] \textit{Quality and Accuracy (including evolution)}. This last aspect has a less specific scope than the previous ones as it addresses the contributions of Ontology to do "better". In practice, it may refer to support advanced functionalities and features or to drive the evolution of a system. A typical practical example emerged from the conducted review is the collaborative approach which, in general terms, requires extended and more sophisticated semantics, as well as a certain level of interoperability. That is, normally, a value provided by ontologies.   

\end{description}

A more extended qualitative analysis of applications aims to estimate the relevance of Ontology in a given context. A summary is reported in Table~\ref{table:relevanceAssessment}. The relevance of knowledge representation, automated reasoning or other features has been informally evaluated without specific criteria independently for each pertinent system to support a more holistic assessment. At least in the 70+\% of cases, the role of ontology may be considered relevant or critical.  

This evaluation is merely indicative and reflects a fundamental lack of detail about the semantic infrastructure and their actual role in many proposed systems.

\begin{table}[!h]
\small
\begin{tabularx}{\columnwidth}{ |X|c|c|c|c| } 
\hline
  \multicolumn{5}{c}{\textbf{Ontology Relevance}}\\
\hline
\textbf{Title} & KR & AR & Other & \textbf{Overall}\\
\hline
\hline
\textit{Supporting Trust in Hybrid Intelligence Systems Using Blockchains}~\cite{fill2020supporting} & $\bullet\bullet\bullet$ & $\bullet\bullet\bullet$ & - & $\bullet\bullet\bullet$\\
\hline
\textit{Hybrid intelligence aspects of programming in* AIDA algorithmic pictures}~\cite{watanobe2014hybrid} & $\bullet\bullet$ & - & - & $\bullet\bullet$\\
\hline
\textit{HIAS: Hybrid Intelligence Approach for Soil Classification and Recommendation of Crops}~\cite{palvannan2023hias} &  $\bullet\bullet\bullet$ & - & $\bullet\bullet\bullet$ & $\bullet\bullet$\\
\hline
\textit{Towards ontology-based intelligent model for intrusion detection and prevention}~\cite{isaza2009towards} & $ \bullet\bullet\bullet$ & $\bullet\bullet\bullet$ & - & $\bullet\bullet\bullet$\\
\hline
\textit{Hybint: a hybrid intelligence system for critical infrastructures protection}~\cite{hingant2018hybint} & - & - & $\bullet$ &  $\bullet$\\
\hline
\textit{Hybrid Intelligence for Driver Assistance}~\cite{hwang2003hybrid} & $\bullet$ & - & - & $\bullet$\\
\hline
\textit{Stargazer: A hybrid intelligence platform for drug target prioritization and digital drug repositioning using streamlit}~\cite{lee2022stargazer} & - & - & $\bullet$ & $\bullet$ \\
\hline
\textit{Explaining Scientific and Technical Emergence Forecasting}~\cite{michaelis2015explaining} & $\bullet$ & - & - & $\bullet$\\
\hline
\textit{Automatic generation of conceptual enterprise models}~\cite{reitemeyer2020automatic} & $\bullet$ & - & - & $\bullet$\\
\hline
\textit{Assessing the impact of automated suggestions on decision making: Domain experts mediate model errors but take less initiative}~\cite{levy2021assessing} & $\bullet\bullet$ & - & - & $\bullet\bullet$\\
\hline
\textit{The noosphere paradigm of the development of science and artificial intelligence}~\cite{palagin2017noosphere} & $\bullet\bullet\bullet$ & - & - & $\bullet\bullet\bullet$\\
\hline
\textit{Combining OWL ontology and schema annotations in metadata management}~\cite{pankowski2011combining} & $\bullet\bullet\bullet$ & - & - & $\bullet\bullet\bullet$\\
\hline
\textit{Human-machine collaboration in online customer service--a long-term feedback-based approach}~\cite{graef2021human} & $\bullet$ & - & $\bullet\bullet\bullet$ & $\bullet\bullet\bullet$\\
\hline
\textit{The text fragment extraction module of the hybrid intelligent information system for analysis of judicial practice of arbitration courts}~\cite{taran2021text} & $\bullet\bullet\bullet$ & - & - & $\bullet\bullet\bullet$\\
\hline
\textit{The hybrid intelligent information system approach as the basis for cognitive architecture}~\cite{chernenkiy2018hybrid} & $\bullet\bullet\bullet$ & - & - & $\bullet\bullet\bullet$\\
\hline
\textit{The Social Web of Things (SWoT)-Structuring an Integrated Social Network for Human, Things and Services}~\cite{cheng2014social} & $\bullet\bullet\bullet$ & - & - & $\bullet\bullet\bullet$\\
\hline
\textit{Similarity measure of agents’ ontologies in cohesive hybrid intelligent multi-agent system}~\cite{listopad2020similarity} & $\bullet\bullet$ & - & $\bullet\bullet\bullet$ & $\bullet\bullet\bullet$\\
\hline
\textit{Context aware ontology-based hybrid intelligent framework for vehicle driver categorization}~\cite{sarwar2022context} & $\bullet\bullet\bullet$ & - &  - &$\bullet\bullet\bullet$\\
\hline
\textit{Hybrid intelligent framework for automated medical learning}~\cite{belhadi2022hybrid}  & $\bullet\bullet\bullet$ & - &  - &$\bullet\bullet\bullet$\\
  \hline
\textit{Agents’ ontologies negotiation in cohesive hybrid intelligent multi-agent systems}~\cite{kirikov2021agents} & $\bullet\bullet$ & $\bullet\bullet$ &  - &$\bullet\bullet\bullet$\\
\hline
\textit{HVAC control via hybrid intelligent systems}~\cite{hadjiski2007hvac} & $\bullet\bullet\bullet$ & - &  - &$\bullet\bullet\bullet$\\
\hline
\textit{Estimating of the similarity of agents’ goals in cohesive hybrid intelligent multi-agent system}~\cite{listopad2020estimating} & $\bullet\bullet$ & - & $\bullet\bullet\bullet$ & $\bullet\bullet\bullet$\\
\hline
\textit{Cohesive hybrid intelligent multi-agent system architecture}~\cite{listopad2020cohesive} & $\bullet$ & - &  -& $\bullet$\\
\hline
\textit{An agent-based hybrid intelligent system for financial investment planning}~\cite{zhang2002agent} & $\bullet$ & - & - & $\bullet$\\
\hline
\textit{Modeling team cohesion using hybrid intelligent multi-agent systems}~\cite{listopad2020modeling} & $\bullet$ & - &  -& $\bullet$\\
\hline
\textit{Ontology support for communicating agents in negotiation processes}~\cite{bravo2005ontology} & $\bullet\bullet\bullet$ & - &  -& $\bullet\bullet\bullet$\\
\hline
\textit{Visualization of team cohesion in hybrid intelligent multi-agent systems}~\cite{rumovskaya2020visualization} & $\bullet$ & - &  -& $\bullet$\\
\hline
\textit{Integration of Knowledge Components in Hybrid Intelligent Control Systems}~\cite{hadjiski2016integration} & $\bullet\bullet\bullet$ & $\bullet\bullet\bullet$ &  -& $\bullet\bullet\bullet$\\
\hline
\textit{Using a CBR approach based on ontologies for recommendation and reuse of knowledge sharing in decision making}~\cite{garrido2008using} & $\bullet\bullet\bullet$ & - & $\bullet\bullet\bullet$& $\bullet\bullet\bullet$ \\
\hline 
\textit{Ontology-guided intelligent data mining assistance: Combining declarative and procedural knowledge}~\cite{charest2006ontology} & $\bullet$ & $\bullet\bullet$ & $\bullet\bullet\bullet$ & $\bullet\bullet$ \\
\hline
\textit{A hybrid fuzzy-ontology based intelligent system to determine level of severity and treatment recommendation for Benign Prostatic Hyperplasia}~\cite{torshizi2014hybrid}& $\bullet\bullet\bullet$ & $\bullet$ & $\bullet\bullet\bullet$ & $\bullet\bullet\bullet$  \\
\hline
\textit{Using hybrid intelligent information system approach for text question generation}~\cite{belyanova2020using} & $\bullet$ & - & - & $\bullet$ \\
\hline
\textit{A Hybrid Intelligent Multiagent System for the Remote Control of Solar Farms}~\cite{hernandez2019hybrid} &$\bullet\bullet\bullet$ & $\bullet\bullet$ & - & $\bullet\bullet\bullet$\\
\hline
\textit{Ontology-based meta-model for hybrid collaborative scheduling}~\cite{varela2020ontology} &$\bullet\bullet\bullet$ & - & - & $\bullet\bullet\bullet$\\
\hline
\textit{An Agent-Based Hybrid Intelligent System for Financial Investment Planning}~\cite{li2010agent} &$\bullet$ & - & - & $\bullet$\\
\hline
\textit{A generic architecture for hybrid intelligent test systems}~\cite{eltaher2008generic} &$\bullet$ & $\bullet\bullet\bullet$ & - & $\bullet\bullet\bullet$\\
\hline
\textit{Constructing hybrid intelligent systems for data mining from agent perspectives}~\cite{zhong2004constructing} &$\bullet$ & - & - & $\bullet$\\
\hline
\textit{Efficient services in the industry 4.0 and intelligent management network}~\cite{martin2017efficient} & $\bullet\bullet\bullet$ & $\bullet\bullet$ & - & $\bullet\bullet\bullet$\\
\hline
\textit{A fog-based hybrid intelligent system for energy saving in smart buildings}~\cite{de2020fog} & $\bullet$ & - & - & $\bullet$\\
\hline
\textit{DSS-based ontology alignment in solid reference system configuration}~\cite{gouveia2020dss} & $\bullet\bullet\bullet$ & - & - & $\bullet\bullet\bullet$\\
\hline
\textit{Design and Conceptual Development of a Novel Hybrid Intelligent Decision Support System Applied towards the Prevention and Early Detection of Forest Fires}~\cite{casal2023design} & $\bullet\bullet$ & - & - & $\bullet\bullet$\\
\hline
\textit{A hybrid reasoning architecture for business intelligence applications}~\cite{krieger2008hybrid} & $\bullet\bullet\bullet$ & $\bullet\bullet$ & - & $\bullet\bullet\bullet$\\

\hline
\hline
\multicolumn{5}{l}{\scriptsize KR = Knowledge Representation, AR = Automatic Reasoning}\\
\multicolumn{5}{l}{\scriptsize ($\bullet\bullet\bullet$) = critical, ($\bullet\bullet$) = relevant, ($\bullet$) = supporting}\\
\hline
\end{tabularx}
\caption{\label{table:relevanceAssessment}Assessment of the relevance of Ontology within the considered system.}
\end{table}

\subsection{Major Research Gaps}

The literature review conducted has also allowed the identification and the consequent formulation of a number of research gaps as follows:

\begin{description}
\item[G.1] The understanding of Hybrid Intelligence in a specific context or system is not always explicitly defined but rather intuitive. This makes hard to understand the effective relevance and role of ontologies within the broader system to solve a given problem or address a given challenge.
\item[G.2] Many contributions address and empathise the added value provided by ontologies. However, there is often a fundamental lack of detail to support the corresponding claims. 
\item[G.3] The contribution of ontologies to achieve hybrid environments where human and artificial intelligence co-exist and co-operate is not fully addressed, although the review has pointed out a relevant insight at a conceptual level.  
\item[G.4] An implicit role of ontologies as an interface between humans and machines in hybrid systems emerged from the conducted literature review. However, looking at existing solutions, the actual potential seems largely unexplored, especially at an application level. 
\item[G.5] Lack of focus on automatic reasoning and inference.
\item[G.6] Lack of focus on ontological modelling. It reflect a fundamental lack of big picture, meaning that within certain systems ontologies are seen within specific components rather than at a more holistic level. 
\end{description}

\subsection{Towards a principled approach}

The HI definition in very generic terms is intuitive and somehow consolidated, even in the context of the current AI mainstreams. A further consolidation step from a conceptual perspective is to establish a reasonable set of principles to characterise hybrid intelligent technology. 

A preliminary critical analysis performed in the context of this review has led to the definition of three principles as follows:

\begin{description}
\item[P.1] Human input is determinant to generate a solution
\item[P.2] Automated solutions are not acceptable solutions to final users
\item[P.3] Rules/conditions to keep the system as hybrid are identified
\end{description}

That same analysis has allowed the identification of a number of open issues at a conceptual level, as reported in Table~\ref{tab:conceptualQuestions}. In the table each question is associated with a relevant aspect, a brief rationale, a link with identified principles and the potential/expected contribution of Ontology. According to this holistic analysis framework, the proposed principles cover mostly the definition of HI in an application context, while Ontology is expected to contribute mainly on engineering aspects by facilitating a better integration between human and machine capabilities, on acceptance through more transparent models and on exploitation by contributing to the generation of dynamic frameworks. On the other side, as briefly discussed in previous sections, an assessment of the evolution of HI in response to advancements in AI is probably unrealistic. 

\begin{table}[H]
\begin{tabularx}{\textwidth}{| X | X  | X | c | c | c}
\hline
\hline
\textbf{Open Question} & \textbf{Aspect} & \textbf{Rationale} & \textbf{Principle} & \textbf{Ontology} \\
\hline
\hline
\textit{When can a system be considered to be Hybrid Intelligent?} & \textbf{\textit{Definition}} & Certain HI-based implementations could be AI solution de facto & P.1, P.2, P.3 & -\\
\hline
\textit{How to fully exploit the potentiality of AI within HI solutions?} & \textbf{\textit{Engineering}} & An effective engineering of HI solutions is a critical issue & - & \checkmark\\
\hline
\textit{Would such a kind of technology be "accepted"?} & \textbf{\textit{Acceptance}} & It could be perceived like a kind of "downgrade" from AI & - & \checkmark\\
\hline
\textit{How to identify critical applications?} & \textbf{\textit{Exploitation}} & Criteria may vary very much from case to case and HI is not always applicable & - & \checkmark\\
\hline
\textit{How will HI evolve?} & \textbf{\textit{Evolution}} &  A more and more advanced AI technology could need a constant re-focus and re-engineering of HI solutions & - & -\\
\hline
\end{tabularx}
\caption{Open questions on HI from a conceptual perspective.}\label{tab:conceptualQuestions}
\end{table}

\section{Conclusions}

In a context of constant evolution and proliferation of AI technology, Hybrid Intelligence is gaining popularity to refer a balanced coexistence between human and artificial intelligence. That same concept has been often used in the recent past to define a model of intelligence resulting from multiple technologies. 

By adopting a soft methodology, this paper proposes a literature review that aims to provide (\textit{i}) a concise and focused overview of the adoption of Ontology in the broad context of Hybrid Intelligence regardless of its definition and (\textit{ii}) a critical discussion on the possible role of Ontology to reduce the gap between human and artificial intelligence within hybrid intelligent systems. 

Beside the typical benefits provided by an effective use of ontologies, at a conceptual level, the analysis conducted has pointed out a significant contribution to quality and accuracy, as well as a more specific role to enable extended interoperability, system engineering and explainable/transparent systems. On the other side, an application-oriented analysis has shown a significant role in present systems (70+\% of the cases) and, potentially, in future systems.

At a more holistic and conceptual level, Ontology is expected to contribute to a better engineering, as well as to contribute to the acceptance and exploitation of hybrid intelligent technology.

As extensively discussed in the paper, the concept of Hybrid Intelligence is evolving to meet the requirements of a new generation of systems that may present blurred boundaries.
A proper holistic discussion on the establishment of the next generation of hybrid-intelligent environments with a balanced co-existence of human and artificial intelligence is fundamentally missed in literature. So is the consequent analysis on the role of ontology.

\end{document}